\documentclass[10pt,a4paper]{article}

\usepackage{graphicx}
\usepackage{amsfonts}
\usepackage{amsmath}
\usepackage{booktabs}
\usepackage{siunitx}
\usepackage{bm}
\usepackage{transparent}
\usepackage{float}
\usepackage{caption}
\usepackage{subcaption}
\usepackage{wrapfig}
\usepackage{ragged2e}
\newcommand{\B}{\fontseries{b}\selectfont} 

\usepackage{natbib}

\newcommand{\power}{power}
\newcommand{\pointcloud}{\mathcal P}
\newcommand{\featurecloud}{\mathcal F}
\newcommand{\pointfeaturecloud}{\pointcloud_\featurecloud}
\newcommand{\kernel}{g}

\newcommand{\npointclouds}{M}
\newcommand{\npoints}{{n_{\pointcloud}}}
\newcommand{\dimpoints}{N}
\newcommand{\numkernelpoints}{K}
\newcommand{\dimfeatures}{D}
\newcommand{\dimstatfeatures}{d}
\newcommand{\dimfeaturesnew}{{D_{\text{out}}}}
\newcommand{\dimfeaturesin}{{D_{\text{in}}}}
\newcommand{\weightmatrix}{W}
\newcommand{\Reals}{\mathbb{R}}
\renewcommand{\vec}[1]{\mathbf{#1}}
\newcommand{\norm}[1]{\left\lVert#1\right\rVert}
\newcommand{\dbh}{\ensuremath{d_{\text{bh}}}}
\newcommand{\ch}{\ensuremath{h_{\text{c}}}}
\newcommand{\dst}{\ensuremath{d_{\text{st}}}}

\usepackage{enumitem} %
\usepackage{color}

\definecolor{turquoise}{cmyk}{0.65,0,0.1,0.3}
\definecolor{purple}{rgb}{0.65,0,0.65}
\definecolor{dark_green}{rgb}{0, 0.5, 0}
\definecolor{orange}{rgb}{0.8, 0.6, 0.2}
\definecolor{red}{rgb}{0.8, 0.2, 0.2}
\definecolor{darkred}{rgb}{0.6, 0.1, 0.05}
\definecolor{blueish}{rgb}{0.0, 0.3, .6}
\definecolor{light_gray}{rgb}{0.7, 0.7, .7}
\definecolor{pink}{rgb}{1, 0, 1}
\definecolor{greyblue}{rgb}{0.25, 0.25, 1}

\newcommand{\real}{\mathbb{R}}

\renewcommand{\paragraph}[1]{\vspace{1em}\noindent\textbf{#1}.}
\usepackage{hyperref}
\hypersetup{pagebackref,breaklinks,colorlinks}

\renewcommand{\th}{$^{\text{th}}$}
\newcommand{\lidar}{\mbox{LiDAR}}
\newcommand{\prm}[1]{} %

\usepackage[capitalize]{cleveref}
\crefname{section}{Sec.}{Secs.}
\Crefname{section}{Section}{Sections}
\Crefname{table}{Table}{Tables}
\crefname{table}{Tab.}{Tabs.}

\pdfoutput=1

\begin{document}
\title{Deep Learning Based 3D Point Cloud Regression for Estimating Forest Biomass}

\author{
\small Stefan Oehmcke\\
\small University of Copenhagen\\
{\tt\small \url{stefan.oehmcke@di.ku.dk}}
\and
\small Lei Li\\
\small University of Copenhagen\\
\small Computer Science\\
{\tt\small \url{lilei@di.ku.dk}}
\and
\small Katerina Trepekli\\
\small University of Copenhagen\\
{\tt\small \url{atr@ign.ku.dk}}
\and
\small Jaime Revenga\\
\small University of Copenhagen\\
{\tt\small \url{jar@ign.ku.dk}}
\and
\small Thomas Nord-Larsen\\
\small University of Copenhagen\\
{\tt\small \url{tnl@ign.ku.dk}}
\and
\small Fabian Gieseke\\
\small University of Münster\\
\small University of Copenhagen\\
{\tt\small \url{fabian.gieseke@di.ku.dk}}
\and
\small Christian Igel\\
\small University of Copenhagen\\
{\tt\small \url{igel@di.ku.dk}}}

\date{}
\maketitle

\begin{abstract}
Quantification of forest biomass stocks and their dynamics is important for implementing effective climate change mitigation measures.
The knowledge is needed, e.g., for local forest management, studying the processes driving \mbox{af-,} re-, and deforestation, and can improve the accuracy of carbon-accounting.
Remote sensing using airborne \lidar{} can be used to perform these measurements of vegetation structure at large scale. 
We present deep learning systems for predicting wood volume, above-ground biomass (AGB), and subsequently above-ground carbon stocks directly from airborne \lidar{} point clouds. 
We devise different neural network architectures for point cloud regression and evaluate them on remote sensing data of areas for which AGB estimates have been obtained from field measurements in the Danish national forest inventory.
Our adaptation of Minkowski convolutional neural networks for regression gave the best results.
The deep neural networks produced significantly more accurate wood volume, AGB, and carbon stock estimates compared to state-of-the-art approaches operating on basic statistics of the point clouds.  
In contrast to other methods, the proposed deep learning approach does not require a digital terrain model.
We expect this finding to have a strong impact on \lidar{}-based analyses of biomass dynamics. 
\end{abstract}

\section{Introduction}\label{sec:intro}

Robust quantification of forest carbon stocks and their dynamics is important for climate change mitigation and adaptation strategies \citep{fao2000}.
The Paris Agreement~\citep{agreement2015paris}
and the IPCC~\citep{shukla2019climate}
acknowledge that climate change mitigation goals cannot be achieved without a substantial contribution from forests. 
Spatial details in the carbon budget of forests are necessary to encourage transformational actions towards a sustainable forest sector~\citep{harris2021global,harris2012baseline}. 
Currently, many countries do not have nationally specific forest carbon accumulation rates but rather rely on default rates from the %
IPCC~2018~\citep{ipcc2018,requena2019estimating}), without accounting for finer-scale variations of carbon stocks \citep{cook2020mapping}.

Precise spatio-temporal monitoring of 
forest carbon dynamics at large scales has proven to be challenging \citep{erb2018unexpectedly,Griscom11645}.
This is due to the complex structure of forests, topographic features, and land management practices~\citep{tubiello2021carbon,lewis2019regenerate}.
Technological developments in remote sensing and the concurrent increased availability of field-based measurements have led to an improvement in estimating carbon stocks using remote sensing observations of forest attributes that serve as proxy for above-ground biomass~(AGB) \citep{knapp2018linking,bouvier2015generalizing,pan2013structure}.

\begin{figure}[t!]
    \centering
    \resizebox{.9\linewidth}{!}{\begingroup%
  \makeatletter%
  \providecommand\color[2][]{%
    \errmessage{(Inkscape) Color is used for the text in Inkscape, but the package 'color.sty' is not loaded}%
    \renewcommand\color[2][]{}%
  }%
  \providecommand\transparent[1]{%
    \errmessage{(Inkscape) Transparency is used (non-zero) for the text in Inkscape, but the package 'transparent.sty' is not loaded}%
    \renewcommand\transparent[1]{}%
  }%
  \providecommand\rotatebox[2]{#2}%
  \newcommand*\fsize{\dimexpr\f@size pt\relax}%
  \newcommand*\lineheight[1]{\fontsize{\fsize}{#1\fsize}\selectfont}%
  \ifx\svgwidth\undefined%
    \setlength{\unitlength}{300bp}%
    \ifx\svgscale\undefined%
      \relax%
    \else%
      \setlength{\unitlength}{\unitlength * \real{\svgscale}}%
    \fi%
  \else%
    \setlength{\unitlength}{\svgwidth}%
  \fi%
  \global\let\svgwidth\undefined%
  \global\let\svgscale\undefined%
  \sffamily
  \makeatother%
  \begin{picture}(1,0.48692104)%
    \lineheight{1}%
    \setlength\tabcolsep{0pt}%
    \put(0,0){\includegraphics[width=\unitlength,page=1]{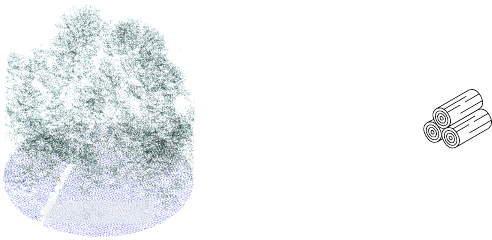}}%
    \put(0.92199738,0.04183183){\color[rgb]{0,0,0}\makebox(0,0)[t]{\lineheight{1.25}\smash{\begin{tabular}[t]{c}AGB\end{tabular}}}}%
    \put(0.58282712,0.04183183){\color[rgb]{0,0,0}\makebox(0,0)[t]{\lineheight{1.25}\smash{\begin{tabular}[t]{c}model\end{tabular}}}}%
    \put(0.20010437,0.04552674){\color[rgb]{0,0,0}\makebox(0,0)[t]{\lineheight{1.25}\smash{\begin{tabular}[t]{c}point cloud\end{tabular}}}}%
    \put(0,0){\includegraphics[width=\unitlength,page=2]{fig/img/pipeline.pdf}}%
  \end{picture}%
\endgroup%
}
  \caption{We use point clouds from airborne \lidar{} to estimate forest biomass and wood volume using deep learning.}
  \label{fig:motivation}
  \end{figure}

Currently, three remote sensing techniques are applied to collect data for AGB estimates: 
i)~passive optical imagery, ii)~synthetic aperture radar~(SAR), and iii)~light detection and ranging~(\lidar{}). 
Compared to \lidar{}, both optical and radar-based data tend to saturate with increasing AGB exceeding 100 \,Mg\,ha$^{-1}$\citep{mermoz2015decrease,sinha2015review,mitchard2012mapping}. 
\lidar{}, as an active sensor, can penetrate dense forest canopies regardless of illumination conditions.
In comparison, SAR observations are sensitive to topographical variations (e.g., steep slopes or cliffs), while imaging spectroradiometer products are highly dependent on the absence of clouds \citep{sinha2015review,treuhaft2014tropical}.
Both alternatives to \lidar{} may not be sufficient to detect variations in AGB in the region of interest \citep{goetz2011advances}.  
Thus, \lidar{} is the better tool to accurately characterize the fine-scale spatial variability in forest structure and terrestrial carbon stocks \citep{zhang2019deep,zolkos2013meta}, but also across broad spatial scales, depending on the platform used (i.e., airborne, drone-based, mobile, and terrestrial laser scanning).  
Recent developments in spaceborne \lidar{}, such as the GEDI mission \citep{coyle2019global}, can be useful for identifying areas with high AGB \citep{lang2021high}. 
However, the coarse spatial resolution of these sensors (sparse \SI{25}{\metre} footprints in case of GEDI~\citep{hancock2019gedi}) makes it impossible to derive fine structural information and to
produce local as well as sufficiently accurate AGB estimates
~\citep{lang2022global,li2020forest,zhang2019deep}. 
Although airborne laser scanning (ALS) is a better-suited technology for monitoring carbon stock variations over countrywide spatial extents, the quality of the forest AGB prediction can vary depending on the applied method and data resolution~\citep{bouvier2015generalizing,magnussen2012fine}.

The state-of-the-art for predicting forest biomass from \lidar{} point clouds is to voxelize the data and compute summarizing statistics of the point distribution along the vertical axis, such as mean heights, relative height quantiles, or metrics of heterogeneity~\citep{magnussen2018lidar,knapp2018linking,d2012estimating,stark2012amazon,mascaro2011evaluating}. 
Together with forest attributes obtained from inventory plots, these simple statistical features then serve as inputs to prediction models~\citep{fassnacht2014importance} such as linear and non-linear regression, random forests, support vector machines, or neural networks, aimed at predicting the forest characteristics of interest.
Based on this approach, several studies are focused on generalizing AGB estimations across different forest types describing different aspects of forest structure at different spatial scales~\citep{bouvier2015generalizing,asner2014mapping,lefsky2002lidar}. 
For instance, \citet{knapp2020structure} assessed a variety of \lidar{} metrics and auxiliary information (e.g., mean canopy height,  maximum possible stand density, maximum possible tree height, vertical canopy heterogeneity, and average wood density) to detect structural forest attributes that may explain stand AGB across different regions. 
However, in this line of research, assessing the optimal combination of \lidar{} metrics that can provide accurate AGB predictions has proven to be a challenging task.

An alternative to manually defining informative features based on  summarizing point clouds statistics is to use deep learning~\citep{lecun2015deep} approaches that directly work on point clouds~\citep{guo2020deep}.
Deep learning on point clouds for regression, however, is an almost unexplored learning setting, except for the estimation of geometric properties from point clouds (e.g., hand pose-estimation~\citealp{ge2018point,jimaging7050080}) and bounding box prediction in general detection tasks~\citep{voxelnet,votenet}. 
In addition, ALS, a technology with peculiarities~\citep{naesset2009effects, goodwin2006assessment}, produces more challenging point clouds than most standard benchmark data sets for point cloud tasks, since they are measured from the top instead of from the ground, and the corresponding point densities are usually lower compared to those induced by other \lidar{}-based technologies (e.g., TLS or drone-based \lidar{}).
The applicability of a deep learning system to estimate crop biomass has also been tested more recently using point clouds from \lidar{} mounted on a roving vehicle ~\citep{pan2022biomass} in a small scale experiment.

This study brings forward deep learning systems to improve forest biomass and wood volume quantification based on airborne \lidar{} data with high spatial resolution per sample (\SI{0.07}{ha}).
Specifically, we propose to apply deep neural networks directly to the point clouds (see \cref{fig:motivation}) instead of relying on models being based on simple statistical features (we presented the initial idea as a  poster in \cite{oehmcke:22}).
We hypothesize that maintaining the full dimensions of the point cloud, and hereby preventing the information loss associated with reducing dimensionality into a few statistical features strengthens predictions. 
The experimental evaluation shows that predictions using deep learning are considerably more accurate than currently employed methods.
Further, in contrast the currently employed methods, our approach requires neither a digital terrain model~(DTM) nor a CHM.
This may pave the way for a paradigm shift for \lidar{}-based analyses of biomass dynamics.

\section{Methods}
In the following, we first introduce the data used in this study in Section~\ref{sec:forest_data}, which combines field-based biomass and wood volume measurements with ALS point clouds. 
Afterwards, in Section~\ref{sec:neural} the assessed deep learning architectures %
are presented along with their required adjustments for point cloud regression. 
The established baseline methods, which rely on statics derived from the point clouds are discussed in Section~\ref{sec:baseline}.
Finally, we give details on why and how to correct the bias for least-square regression tasks in Section~\ref{sec:bias}.

\subsection{Forest Data}\label{sec:forest_data}
This section gives details on the two used data sources and their processing. 
First, the inventory data of AGB and wood volume are described, as well as how they were collected.
Then, we report on the \lidar{} data linked to these measurements.

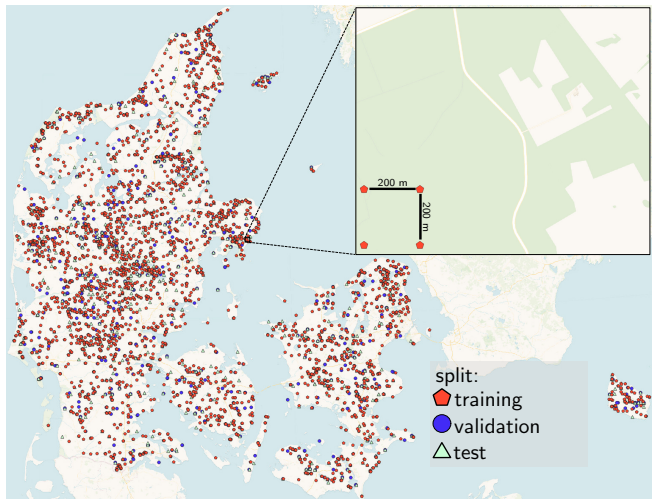
\begin{figure}
    \centering
    \resizebox{.7\linewidth}{!}{\begingroup%
  \makeatletter%
  \providecommand\color[2][]{%
    \errmessage{(Inkscape) Color is used for the text in Inkscape, but the package 'color.sty' is not loaded}%
    \renewcommand\color[2][]{}%
  }%
  \providecommand\transparent[1]{%
    \errmessage{(Inkscape) Transparency is used (non-zero) for the text in Inkscape, but the package 'transparent.sty' is not loaded}%
    \renewcommand\transparent[1]{}%
  }%
  \providecommand\rotatebox[2]{#2}%
  \newcommand*\fsize{\dimexpr\f@size pt\relax}%
  \newcommand*\lineheight[1]{\fontsize{\fsize}{#1\fsize}\selectfont}%
  \ifx\svgwidth\undefined%
    \setlength{\unitlength}{300bp}%
    \ifx\svgscale\undefined%
      \relax%
    \else%
      \setlength{\unitlength}{\unitlength * \real{\svgscale}}%
    \fi%
  \else%
    \setlength{\unitlength}{\svgwidth}%
  \fi%
  \global\let\svgwidth\undefined%
  \global\let\svgscale\undefined%
  \sffamily%
  \makeatother%
  \begin{picture}(1,0.77049192)%
    \lineheight{1}%
    \setlength\tabcolsep{0pt}%
    \put(0,0){\includegraphics[width=\unitlength,page=1]{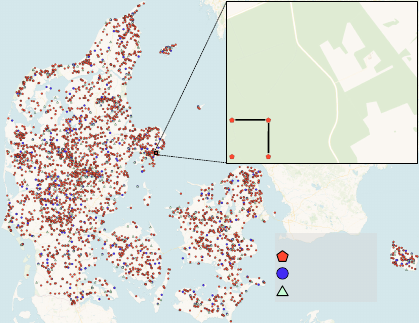}}%
    \put(0.66504166,0.18244275){\makebox(0,0)[lt]{\lineheight{1.25}\smash{\begin{tabular}[t]{l}split:\end{tabular}}}}%
    \put(0.69377766,0.1452033){\makebox(0,0)[lt]{\lineheight{1.25}\smash{\begin{tabular}[t]{l}training\end{tabular}}}}%
    \put(0.69370967,0.10240332){\makebox(0,0)[lt]{\lineheight{1.25}\smash{\begin{tabular}[t]{l}validation\end{tabular}}}}%
    \put(0.69239809,0.0630694){\makebox(0,0)[lt]{\lineheight{1.25}\smash{\begin{tabular}[t]{l}test\end{tabular}}}}%
    \put(0.5985346,0.48858503){\makebox(0,0)[t]{\lineheight{1.25}\smash{\begin{tabular}[t]{c}{\tiny200 m}\end{tabular}}}}%
    \put(0.64539277,0.44060635){\rotatebox{-90}{\makebox(0,0)[t]{\lineheight{1.25}\smash{\begin{tabular}[t]{c}\tiny200 m\end{tabular}}}}}%
  \end{picture}%
\endgroup%
}
    \caption{Measurement sites (Danish NFI plots). Color indicates how the data were split into training, validation, and testing 
    (map background based on \cite{OpenMapTiles} and \cite{CARTO}).
    }
    \label{fig:map}
\end{figure}

\subsubsection{Forest Inventory}
\begin{figure}
\begin{subfigure}[b]{.49\linewidth}
    \centering
    \resizebox{\linewidth}{!}{\input{fig/hist_targets_}}
    \caption{Histogram of above-ground forest biomass, carbon stock, and wood volume.}
    \label{fig:target_hist}
\end{subfigure}
\begin{subfigure}[b]{.49\linewidth}
\centering
\resizebox{.75\linewidth}{!}{\begingroup%
  \makeatletter%
  \providecommand\color[2][]{%
    \errmessage{(Inkscape) Color is used for the text in Inkscape, but the package 'color.sty' is not loaded}%
    \renewcommand\color[2][]{}%
  }%
  \providecommand\transparent[1]{%
    \errmessage{(Inkscape) Transparency is used (non-zero) for the text in Inkscape, but the package 'transparent.sty' is not loaded}%
    \renewcommand\transparent[1]{}%
  }%
  \providecommand\rotatebox[2]{#2}%
  \newcommand*\fsize{\dimexpr\f@size pt\relax}%
  \newcommand*\lineheight[1]{\fontsize{\fsize}{#1\fsize}\selectfont}%
  \ifx\svgwidth\undefined%
    \setlength{\unitlength}{130bp}%
    \ifx\svgscale\undefined%
      \relax%
    \else%
      \setlength{\unitlength}{\unitlength * \real{\svgscale}}%
    \fi%
  \else%
    \setlength{\unitlength}{\svgwidth}%
  \fi%
  \global\let\svgwidth\undefined%
  \global\let\svgscale\undefined%
  \makeatother%
  \sffamily%
  \textbf{
  \begin{picture}(1,0.99587624)%
    \lineheight{1}%
    \setlength\tabcolsep{0pt}%
    \put(0,0){\includegraphics[width=\unitlength,page=1]{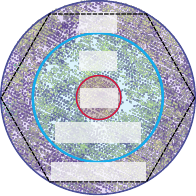}}%
    \put(0.50288921,0.46798058){\color[rgb]{0,0,0}\makebox(0,0)[t]{\lineheight{1.25}\smash{\begin{tabular}[t]{c}\SI{3.5}{m}\end{tabular}}}}%
    \put(0.50288824,0.65379887){\color[rgb]{0,0,0}\makebox(0,0)[t]{\lineheight{1.25}\smash{\begin{tabular}[t]{c}\SI{10}{m}\end{tabular}}}}%
    \put(0.50288824,0.85200106){\color[rgb]{0,0,0}\makebox(0,0)[t]{\lineheight{1.25}\smash{\begin{tabular}[t]{c}\SI{15}{m}\end{tabular}}}}%
    \put(0.49999995,0.295){\color[rgb]{0,0,0}\makebox(0,0)[t]{\lineheight{1.25}\smash{\begin{tabular}[t]{c}{\small$\dbh{} > \SI{10}{\centi\metre}$}\end{tabular}}}}%
    \put(0.49999995,0.09){\color[rgb]{0,0,0}\makebox(0,0)[t]{\lineheight{1.25}\smash{\begin{tabular}[t]{c}{\small$\dbh{} > \SI{40}{\centi\metre}$}\end{tabular}}}}%
  \end{picture}%
  }
\endgroup%
}%
\caption{Subplot measurement circle. }\label{fig:circle}%
\end{subfigure}%
\caption{%
The left plot (a) shows the distribution of target variables. 
Above-ground biomass and carbon stocks %
    overlap completely since the carbon stock estimate is computed as a linear function of the biomass. %
The right plot (b) shows exemplary input point cloud (top view of a single subplot) and a representation of the measurement circles. The dashed hexagon encloses the points used by the deep learning point cloud methods.
}
\end{figure}

The volume (in m$^3$\,ha$^{-1}$)  and biomass (Mg (metric ton) ha$^{-1}$) estimates used to train and evaluate the deep learning models stem from the Danish National Forest Inventory~(NFI), see \cref{fig:target_hist} for their distribution.
The Danish NFI is based on a grid with cells of size $2 \times \SI{2}{\kilo\metre}$ covering the entire land surface of the country~\citep{nord2016danish}. 
A plot is composed of four circular subplots with a radius of \SI{15}{\metre}.
Each subplot is located in the corners of a $200 \times \SI{200}{\metre}$ square, which is randomly placed within each grid cell (\cref{fig:map}).
The full set of plots is geographically partitioned into five spatially balanced interpenetrating panels~\citep{kish1998space,mcdonald2003review,olsen1999statistical,zhang2003smoothing}.
Each year within a 5-year cycle, a different panel is measured, representing one fifth of the total set but representative of the entire country.
These labeled data used in our study range from 2013 to 2017.

In the Danish NFI, each subplot is composed of three concentric circles with radii of \SI{3.5}{\metre}, \SI{10}{\metre}, and \SI{15}{\metre}, respectively.
Following standard procedures, a single caliper measurement of diameter is made at breast height (\dbh) (i.e., \SI{1.3}{\metre} from the ground) for all trees in the \SI{3.5}{\metre} circle.
Trees with \dbh{} larger than \SI{10}{\centi\metre} are measured in the \SI{10}{\metre} circle, and only trees with \dbh{} larger than \SI{40}{\centi\metre} were recorded in the \SI{15}{\metre} circle (see \cref{fig:circle}).
For a random sub-sample, further measurements of the total height (\textit{h}), crown height (\ch), age, and diameter at stump height~(\dst) of two to six trees within each subplot are also obtained.

Based on the sub-sample measured for both $h$ and \dbh, models were developed for each species and growth region. 
These models are based on the observed mean height and mean diameter within each subplot for creating localized regressions using the approach suggested by \cite{sloboda1993regionale}.
For subplots where no height measurements were available, generalized regressions were developed based on a modified  Näslund-equation~\citep{naslund1936skogsforsoksanstaltens,Johannsen1999}.
Subsequently, individual tree volume and biomass were estimated using %
species-specific models~\citep{NordLarsenetal2017}.
Since trees were measured in different concentric circles depending on their diameter, the volume of growing stock and biomass in each subplot were calculated by scaling the estimated AGB and wood volume of each tree according to the circular area in which the tree was measured.
Carbon stocks can be considered as a linear function of the forest biomass, thus, in this study was not used as a separate regression target.

\begin{figure}
    \centering
    \begin{subfigure}[b]{\linewidth}
        \includegraphics[width=\linewidth]{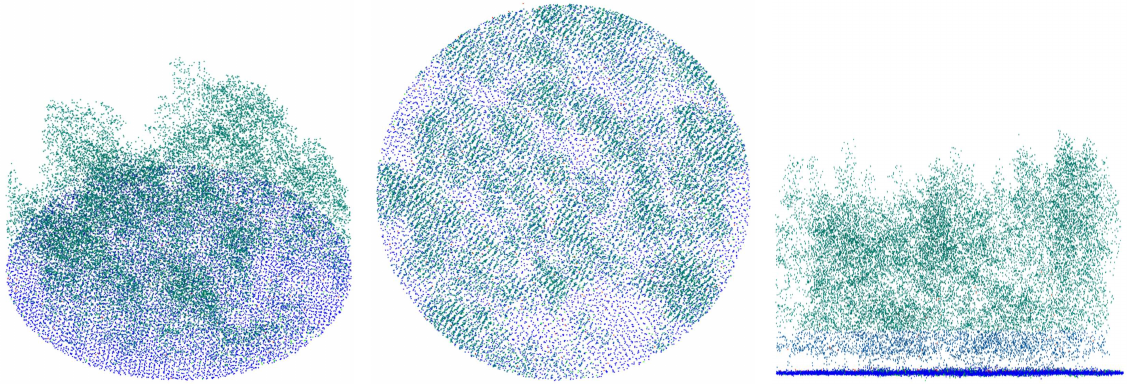}
    \caption{Broadleaf only subplot with AGB of \num{27.93} \,Mg\,ha$^{-1}$.}
    \end{subfigure}
    
    \begin{subfigure}[b]{\linewidth}
    \includegraphics[width=\linewidth]{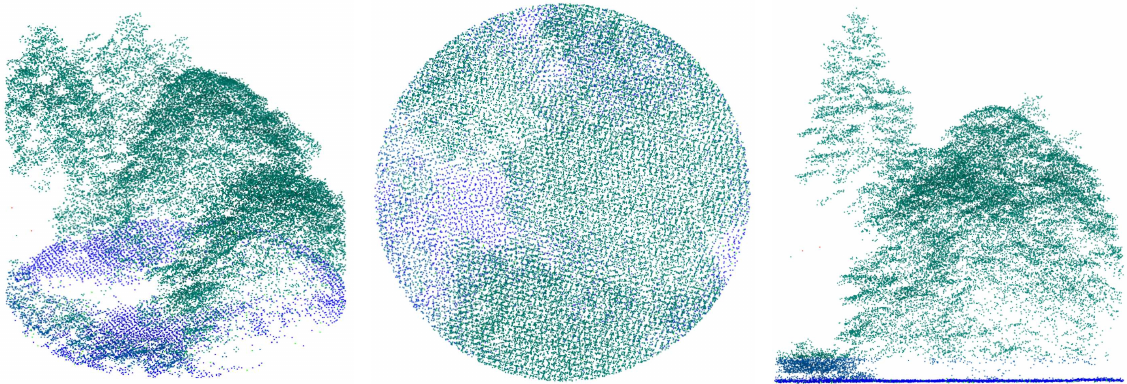}
    \caption{Mixed forest subplot (28\% conifer) with AGB of \num{226.50} \,Mg\,ha$^{-1}$.
    }
    \end{subfigure}

     \begin{subfigure}[b]{\linewidth}
        \includegraphics[width=\linewidth]{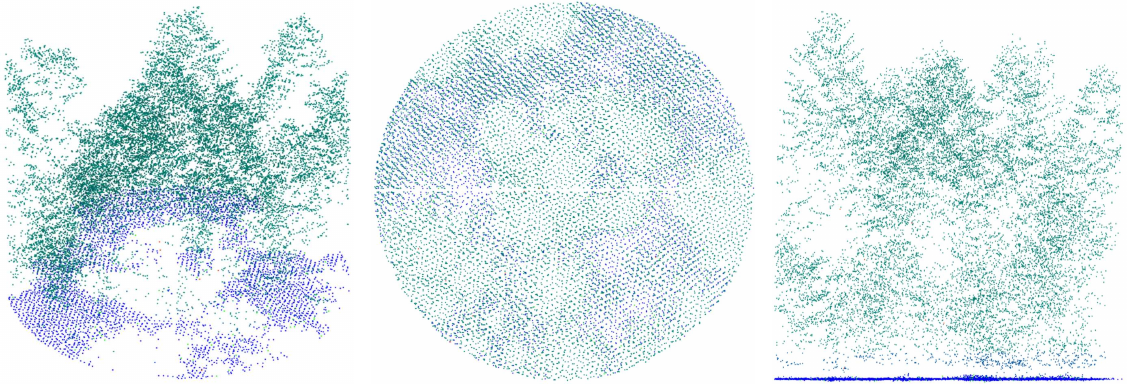}
    \caption{Conifer only subplot with AGB of \num{150.59} \,Mg\,ha$^{-1}$.}
    \end{subfigure}
    \begin{subfigure}[b]{\linewidth}
        \includegraphics[width=\linewidth]{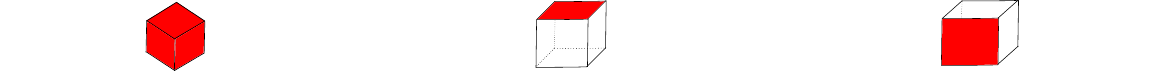}
    \end{subfigure}

    \caption{
    Examples of subplots with different fractions of broadleaf and conifer trees. Green and blue indicate points classified belonging to trees and ground, respectively. We show three perspectives: isometric front, top, and side view.}
    \label{fig:samples}
\end{figure}

\subsubsection{Point Cloud Data} 
Airborne \lidar{} point cloud data covering the whole of Denmark from 2014 to 2018 are publicly available from the Danish \emph{Agency for Data Supply and Efficiency} (\url{https://dataforsyningen.dk}).
Specifically, a sampling campaign in 2014 and 2015 covered the entire country, while a fifth of the country was again scanned in 2018 (in eastern Jutland and Funen).
The average resolution of the point clouds was reported to be \num{4.5} points per m$^2$ 
with a vertical and horizontal point accuracy 
 (given as root mean squared error~(RMSE)) of \SI{5}{\centi\metre} and \SI{15}{\centi\metre}, respectively.
We extracted the point clouds corresponding to the subplots studied in the NFI (see \cref{fig:samples}).
The average number of points per subplot equaled \num{11684} with a standard deviation of \num{6743} points.

\subsubsection{Preprocessing}
Several preprocessing steps were conducted for both the deep neural networks (operating on the point clouds) as well as for the baseline models that operate on simple statistical features. 

For the plot-specific point clouds, the NFI identified four types of errors.
In the first type, trees visible in the point cloud  do not belong to the field-based measurements but reach into the plot from the outside, while in the second type non-forest objects, such as land-lines or buildings, are part of the point cloud.
We chosen not to exclude these data since a method that directly works with the point cloud should detect irrelevant parts by itself, meaning that some previously excluded data should now be usable.
The third error type occurs when the plot had been harvested in-between field measurements and airborne point cloud acquisition.
Finally, the fourth error type consisted of unreasonable values. 
Since the third and fourth error flags are intractable, the samples were excluded from the model training.
In addition, all samples with no points above \SI{1.3}{\metre} were removed, as this was the minimal height considered for the biomass measurements\footnote{Keeping these samples would simply reduce the errors, because they could be handled by a simple rule that predicts zero if the maximum height is below \SI{1.3}{\metre}.}.

The remaining dataset consists of $\npointclouds=6101$ individual point clouds, each being labelled with a corresponding biomass and wood volume measurement as described in \cref{sec:forest_data}.
The biomass and wood volume were  measured at different times than the collection of point clouds.
To ensure that both data sources matched, we only considered point clouds observed within one year of the biomass measurement for our validation and test set. 
For training, we took larger temporal gaps (up to nine years) into account and added the time interval in years as a feature to the point clouds.
Since some sites were measured multiple times, we kept measurements from the same site together in either training, validation, or test datasets (i.e., no site/plot considered during training or for validation occurred in the test set).

Finally, to train, validate and test all the models, we split the data as follows: 
the training set contains \num{3232} samples with a time interval of more than one year and \num{1039} samples with an even longer time interval (i.e.,  \num{4271} training samples in total).
The validation set consists of \num{919} and the test sets of \num{911} samples with a maximum distance of one year between paired point cloud and biomass measurements.

\subsection{Deep Regression for Point Clouds}\label{sec:neural}

From a mathematical perspective, a single point cloud can be represented via a set
$\pointcloud = \{\vec{x}_1,\ldots,\vec{x}_\npoints\} \subset \Reals^N$,
where~$\npoints$ is the number and $\dimpoints$ the dimensionality of the points (typically, $\dimpoints=3$). 
Besides these coordinates, additional information is often provided for each point, such as color or intensity. 
These features paired with the points define the set
\begin{equation}
\pointfeaturecloud = \{(\vec{x}_1, \vec{f}_1), \ldots, (\vec{x}_\npoints, \vec{f}_\npoints)\} \subset \Reals^\dimpoints \times \Reals^\dimfeatures   
\end{equation}
with $\vec{f}_i \in \Reals^\dimfeatures$ containing the additional features. 
Such a point cloud is given for each instance (e.g., one subplot corresponds to one point cloud, see \cref{fig:samples}).
The point cloud sets typically have different cardinalities and are unordered.
Accordingly, the output of a model should not depend on the order of the provided points, a property referred to as permutation invariance. 
Furthermore, the spatial structure of such point clouds can be highly irregular and the density of points is typically very sparse (such characteristics usually vary among different regions as well).

Compared to deep convolutional neural networks~(CNNs) for standard 2D and 3D images,
the research field of deep learning for point cloud data has yet to mature.
There are several approaches for adapting neural networks for point cloud processing,
differing among others in how they deal with the key questions of how to extend the concept of spatial convolutions to sparse point clouds.
Most deep learning systems for point clouds are developed for segmentation and other classification tasks~\citep{semantic3d,shapenet,modelnet}. 
In this study, we focused on regression tasks. 
We adapted and compared three widely-used and conceptually different point cloud classification approaches for regression: the classic PointNet~\citep{qi2017cvpr},  the kernel point convolution (KPConv)~\citep{Thomas2019}, and
a Minkowski CNN~\citep{choy20194d}.

We extended the open-source library Torch-Points3D~\citep{tp3d} to support regression and will contribute our networks and anonymized dataset to the library upon acceptance of this manuscript.

\subsubsection{PointNet}
The seminal PointNet is one of the first deep learning architectures developed for point clouds~\citep{qi2017cvpr}.
PointNet-type networks apply the same weight matrices to each input point. 
That is, the resulting feature maps are defined by a shared (dense) neural network. 
The original architecture is based on the use of multiple such neural network blocks. 
In between these blocks, input/feature transformations are used to align the data via learnable transformation matrices. 
At the end, to facilitate a variable number of points and to achieve permutation invariance, a symmetric set function is applied, which can be simply a global max-pooling operation. 
Two PointNet variants have been proposed in the seminal work to address classification and segmentation scenarios, respectively.

To adapt the architecture for regression, we adopted the classification PointNet architecture up to the global feature aggregation. 
We then replaced the subsequent two hidden fully-connected layers (output dimensionalities: 512-256) with three fully-connected layers (output dimensionalities: 512-256-128) and considered a final linear output layer.

\subsubsection{KPConv}
The {kernel point convolution}~\citep{Thomas2019} extends standard discrete convolution to the domain of point clouds. 
Kernel point convolutions consider, for a given input point $\vec{x} \in \Reals^\dimpoints$, a local neighborhood along with points called kernel points that are used to define a convolution operator.
More precisely, $\pointcloud_\featurecloud$ is convolved by a kernel~$\kernel$ at a point $\vec{x} \in \Reals^\dimpoints$ is defined via

\begin{equation}
\label{eq:kpconvolution}
\left( \pointfeaturecloud * \kernel \right) (\vec{x}) = \sum_{(\vec{x}_i,\vec{f}_i) \in \mathcal{N}_{\vec{x}}} \kernel(\vec{x}_i - \vec{x}) \vec{f}_i    \enspace.
\end{equation}

\noindent Here $\mathcal{N}_{\vec{x}} = \{(\vec{x}_i,\vec{f}_i) \in \pointfeaturecloud \mid \norm{\vec{x}_i - \vec{x}} \leq r\}$, where $r\in \Reals^+$ is a hyper-parameter defining the domain
$\mathcal{B}_r^\dimpoints=\{\vec{y} \in \Reals^\dimpoints \mid \norm{\vec{y}} \leq r\}$
of the kernel $\kernel$.
The kernel $\kernel$ is based on a set $\{\widetilde{\vec{x}}_1,\ldots,\widetilde{\vec{x}}_\numkernelpoints \} \subset \mathcal{B}_r^\dimpoints$ of $\numkernelpoints$ {kernel points}, where each kernel point $\widetilde{\vec{x}}_k$ has an associated weight matrix $\weightmatrix_k \in \Reals^{\dimfeaturesin \times \dimfeaturesnew}$, which maps a feature vector $\vec{f} \in \Reals^\dimfeaturesin$ to a new feature vector $\vec{f}' \in \Reals^\dimfeaturesnew$ (for the first such convolution layer, we have $\dimfeaturesin = \dimfeatures$). 
One option for the kernel $g:\mathcal{B}_r^\dimpoints \rightarrow \Reals^{\dimfeaturesin \times \dimfeaturesnew}$ is 

\begin{equation}
    g(\vec{y}) = \sum_{k=1}^\numkernelpoints h(\vec{y}, \widetilde{\vec{x}}_k) \weightmatrix_k  \enspace,
\end{equation}

\noindent where $h=\max(0,1-\frac{\norm{\vec{y} - \widetilde{\vec{x}}_k}}{\sigma})$ and  $\sigma \in \Reals^+$ is a user-defined model parameter~\citep{Thomas2019}.
Thus, the kernel yields a weighted sum of weight matrices and each kernel is parameterized by its own kernel points and the corresponding weight matrices. 
The weight matrices are learnt similarly to the coefficients of convolution operators in standard CNNs.
The kernel points can either also be learnt (``deformable'') or can be fixed (``rigid'') systematically around the center of $\mathcal{B}_r^\dimpoints$~\citep{Thomas2019}.
Pooling operations are realized via creating a 3D grid in which all points in a cell are aggregated (see \cref{sec:Minkowski}).

We adapt the KPConv architecture proposed for classification (KP-CNN) to utilize it for regression.
KP-CNN follows the structure of ResNet, thus modifying it for regression was straightforward and was achieved by changing the output layer to have the identity as activation function.
We resorted to fixed kernel points since deformable ones consistently performed worse on our dataset.

\subsubsection{Minkowski Convolutional Neural Network}\label{sec:Minkowski}
A straightforward approach to process point clouds is to voxelize the points to a 3D image (with $D$ channels) and to apply a standard 3D convolutional neural network (CNN) to the resulting image.
This requires binning, but while a coarse binning may discard important information, a fine-grained binning can render the computations intractable. 
However, the points are typically very sparse in 3D space. 
This lends to keeping a sparse representation and applying spatially sparse 3D convolutions, which basically  operate only in areas where points exist~\citep{grahamsparse,graham20183d}.
A state-of-the-art implementation of sparse convolutions for point clouds is given by the Minkowski Engine, an auto-differentiation library, which allows to efficiently implement standard architectures, such as ResNet with 3D convolutions~\citep{choy20194d}.

Both the Minkowski and KPConv architectures progressively downsample the point cloud based on a 3D grid via pooling operations.
The initial grid size is only bound to the horizontal and vertical accuracy of the point data, meaning that a too small grid size cannot be processed efficiently.
Therefore, it is crucial to set the initial grid size according to the data and task.
The subsequent pooling steps are  then based on this initial size and work similarly to standard 2D pooling operations.
Once a grid is created, a voxel-based approach, such as Minkowski CNNs, aggregates all voxels within a grid cell to a new voxel.
In contrast, KPConv takes the mean of the position of all points within a cell to form a new point. %
By doubling the size of the grid cells in each subsampling/pooling layer, the number of points/voxels is gradually reduced, corresponding to an increase in the size of the receptive field
The features associated with each new point can be obtained, for example, via max-pooling (applied to the feature vectors of the pooled points) or by applying the chosen convolution operator. %

\begin{figure}
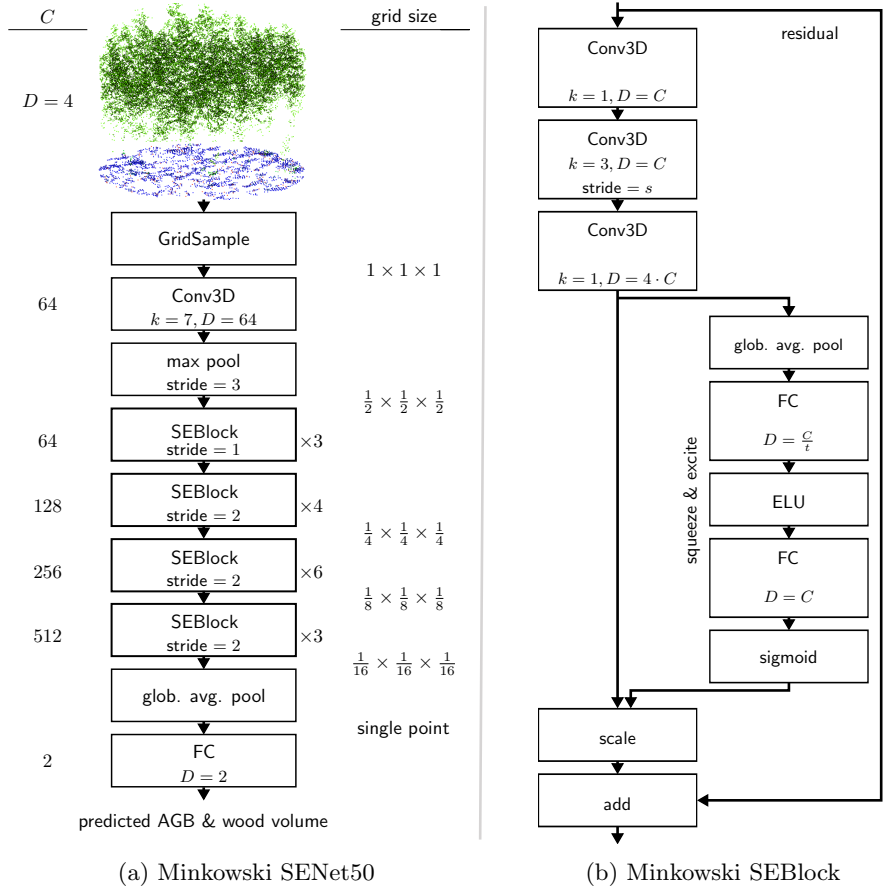

    \centering
    \begin{subfigure}[b]{0.579\linewidth}
    \centering
    \resizebox{.9\linewidth}{!}{\input{fig/img/MSENet_}}
    \caption{Minkowski SENet50}
    \end{subfigure}~%
    \begin{subfigure}[b]{.421\linewidth}
    \centering
    \resizebox{.9\linewidth}{!}{\input{fig/img/MSEBlock_}}
    \caption{Minkowski SEBlock}
    \end{subfigure}
    
    \caption{(a)
    Implemented Minkowski SENet50 (MSENet50) architecture. 
    The grid sizes are relative to the initial grid size.
    Conv3D refers to sparse 3D convolution with $\dimfeatures$ channels followed by ELU activation and batch normalization. The kernel size $k$ and stride are given as single numbers but extend to all three dimensions (e.g., $3 = 3\times3\times3$). FC layers are fully connected layers of dimensionality $\dimfeatures$. (b) Bottleneck block, parameterized by its input size $C$, stride $s$, and reduction rate $t$ (defaults to 16).}
    \label{fig:SENet}
\end{figure}

Using the Minkowski Engine library, we built two 3D versions of SENet (SENet14 and SENet50) for regression \citep{hu2018senet}, see \cref{fig:SENet}.
To that end, we replaced 2D convolutions with their 3D equivalent.
The stride of the initial convolution of SENet was changed from \num{2} to \num{1} in order to avoid early loss of information.
We also replaced Rectified Linear Unit~(ReLU) activation functions with Exponential Linear Units~(ELU)~\citep{ELU}.
Finally, the output layer was changed to fit the number of target variables and does not apply a non-linear activation. 
The two network versions differ in the number of trainable parameters as shown in  \cref{tab:params_time} and are referred to as MSENet14 and MSENet50, respectively.

\subsubsection{Experimental Setup for Forest Attribute Regression }
Three different neural network architectures were applied: the proposed regression versions of PointNet, KPConv, and two 3D version of SENet (MSENet14 and MSENet50) built with the Minkowski Engine.
All code was written in Python and PyTorch~\citep{pytorch} and trained on A100 GPUs.
The positions of the input points were given in the projected coordinate system EPSG:25832.
We centered the $x$-, $y$-, and $z$-coordinates for each subplot individually to get a local reference frame.
Scaling was applied by dividing by \num{15}, \num{15}, \SI{20}{\metre}, respectively, so that the coordinates lie within \([-1, 1]\) for the $x$- and $y$-coordinates.
We do not need to calibrate our model using a DTM, as is required by state-of-the-art methods based on point cloud statistics.

We tuned the initial grid size of KPConv and the MSENet models by gradually decreasing it (i.e., increasing the spatial resolution) until no improvement could be seen on the validation set.
Both architectures worked best with the initial grid size set to \num{0.025}.
The number of trainable model parameters and runtimes are given in \cref{tab:params_time}.
\begin{table}[t]
    
    \centering
    \caption{Number of trainable parameters  and average runtime (elapsed real time using a single GPU) for one pass through the test set as well as the total training time for the point cloud models.}
    \begin{tabular}{lS[table-format=8]rrr}
    \toprule
        Model & \multicolumn{1}{r}{number of parameters} & test time & train time \\ \midrule
        PointNet & 3503691 & 7s & 7h \\
        KPConv & 12287232 & 55s & 42h \\ 
        MSENet14 & 14403966 & 4s & 4h \\ 
        MSENet50 & 48716722 & 5s & 9h \\ 
        
    \bottomrule
    \end{tabular}
    \label{tab:params_time}
\end{table}

To increase diversity in the training set, several augmentations were considered.
First, a random rotation around the z-axis was applied.
We did not consider rotating the x- and y-axis since a tilt would result in unrealistic samples (e.g., change of growth direction).
Thereafter, random dropout of points with a dropout rate of 20\% in 50\% of all samples was applied, which increases variety and simulates changes in point density.
Finally, each dimension of the point position was shifted by adding a small random value drawn from a zero-mean truncated Gaussian distribution with variance \num{0.001} and support $[-0.05,05]$.

One goal was to create precise wood volume and above-ground biomass maps over large areas.
Therefore, instead of using the entire circle in which the measurements were taken, we resorted to a hexagonal shape in the x- and y-coordinate, see \cref{fig:circle}.
This approach facilitates an easy tiling of large areas without gaps and overlaps.
The hexagonal shape covers almost all of the points, in particular those in the inner circles (in control experiments, we verified that considering the whole subplot does not lead to a significant performance increase).

We employed the AdamW~\citep{adamw} optimizer with weight decay set to \num{0.01}.
The initial learning rate was \num{0.001}, which was adapted according to the cosine annealing schedule with warm restarts (\(T_0=10, T_{\text{mult}}=2\))~\citep{cosineawr} for \num{310} epochs. 
The batch size was \num{32} and the smooth L1 function \cite{girshick2015fast} was used as training loss.
These hyper-parameter configurations were chosen based on the highest $R^2$ score on the validation set.
The input channel was filled with ones to encode the presence of a point.

\subsection{Baseline Methods}\label{sec:baseline}

The baseline models considered in this study for estimating above-ground biomass, rely on precomputed features
 from the normalized height distribution in accordance with \cite{nord2017}: 
mean height, %
height standard deviation,
coefficient of variation,
skewness, 
kurtosis, 
percentiles of the height distribution (5\th, 10\th, 25\th, 50\th, 75\th, 90\th, 95\th, 99\th), and the interception ratio (IR), which is the fraction of points above one meter compared to all first returns.
For each point cloud sample, these features except the IR were computed twice, once for all points and once only for the points \SI{1}{\metre} above ground.
To determine the height of a point relative to the ground, we calibrated it semi-automatically, using two DTMs with different spatial resolutions (\SI{25}{\metre^2} and \SI{5}{\metre^2}), and  the first return of each light pulse from the airborne \lidar{}, following the study of \cite{magnussen2018lidar}. 
In preliminary analysis, it was observed that this approach led to better results than using all the recorded returns from each \lidar{}-derived light pulse. 
In contrast, the deep learning methods simply use all the information included in the point clouds without the need to filter out the point cloud datasets according to the number of returns.
In total, $\dimstatfeatures=28$ statistical features were extracted for each point cloud sample.

\subsubsection{Linear and Power Regression}

The first two baseline methods were linear regression and power regression models. 
For the linear regression models, all the $\dimstatfeatures$ features were used as explanatory variables. 
The multivariate \power{} regression can be regarded as the state-of-the art for predicting AGB, above-ground carbon stocks, and tree volume from the point cloud features~\citep{magnussen2018lidar}.
Following \cite{magnussen2018lidar}, we  fitted models of the form
\begin{equation}
    y=w_1 \cdot z_{\text{mean AG}}^{w_2} \cdot z_{95^{\text{th}} \text{AG}}^{w_r} \cdot \text{IR}^{w_4}
\end{equation}
for each quantity of interest. 
Here, $z_{\text{mean AG}}$ is the mean height of the points one meter above ground, $z_{95^{\text{th}} \text{AG}}$ is the 95\th\ percentile of the height distribution of point one meter above ground, and IR is the inception ratio.
For each model, the parameters $w_1,\ldots,w_4 \in \Reals$ were set to the optimal least-squares estimates.

\subsubsection{Random Forest}

We trained random forest (RF) models with \num{5000} trees~\citep{breiman2001random} and resorted to the popular Scikit-Learn implementation~\citep{pedregosa2011scikit}.
The out-of-bag (OOB) error was used to tune the RF hyper-parameters.
We considered
$\{0.1, 0.2, \ldots, 1\}$ for the ratio of features to consider at every split,
$\{0.1, 0.2, \ldots, 1\}$ for the ratio of samples to consider for each tree,
$\{5, 6, \ldots, 20, \infty\}$ for the maximum tree depth, and
$\{1, 2, 4, \ldots,16\}$ for the minimum number of samples required at each leaf node.
The best model used 90\% of the features at every split, 20\% of the training data for each tree, had a maximum depth of \num{11}, and required a 
minimum number of \num{6} samples at each leaf node.

\subsection{Bias Correction}

\label{sec:bias}
Least-squares regression with deep learning models typically  lead to biased models in the sense that the sum of residuals on the training data set is not zero \citep{correctBias}. 
 Consider a regression model $\mathbb R^d\to\mathbb R$ of the form 
 
 \begin{equation}\label{eq:wrap}
    f(\vec{x}) = \vec{a}^{\text{T}} h_{\vec{\theta}}(\vec{x})+b   %
\end{equation}

\noindent with parameters $(\vec{\theta}, \vec{a}, b)$. Here $h_{\vec\theta}:\mathbb R^d\to\mathbb R^m$ for some positive integer $m$,  $b\in\mathbb R$, $\vec{a}\in\mathbb R^m$, $\vec{x} \in \mathbb R^d$.
This model can be a deep neural network, where all layers but the final layer are represented by the function $h_{\vec\theta}$ and the parameters $\vec{\theta}$ comprise all weights except those in the final layer. 
Training a deep neural network for regression using iterative optimization with mini-batches and stopping after a fixed training time or based on performance on a hold-out data set cannot be expected to chose the parameter $b$ in a way that the residuals on the training data sum to zero. 
That is, on the training data $\{(\vec{x}_1,y_1),\dots(\vec{x}_N,y_N)\}$ we have 
$\sum_{i=1}^N (y_i -  (\vec{a}^{\text{T}} h_{\vec{\theta}}(\vec{x}_i)+b)) \neq 0$.
This can introduce a systematic error that accumulates if we are interested in the total aggregated performance over many data points. 
We therefore used the method proposed by \cite{correctBias} to correct this bias by replacing $b$ by  

\begin{equation}\label{eq:bstar}
b^* = 
    \frac{\sum_{i=1}^N (y_i -  \vec{a}^{\text{T}} h_{\vec{\theta}}(\vec{x}_i))}{N} \enspace.
\end{equation}

\noindent This bias correction can also be applied to other non-linear models $h_{\vec\theta}$ (e.g.,  $h_{\vec\theta}$ could a random forest); $h_{\vec\theta}$ can be wrapped according to \eqref{eq:wrap} with scalar $a=1$ and $b$ set to $b^*$ computed by \eqref{eq:bstar}.

\section{Results}\label{sec:results} %
We conducted a detailed comparison of all methods for predicting AGB (and thereby above-ground carbon stocks)
and wood volume. We used the root-mean-square-error~(RMSE), the coefficient of determination $R^2$, and the mean absolute percentage error~(MAPE) as evaluation metrics, in accordance with the related literature.
The prediction bias parameter was calibrated on the training and validation set for all non-linear methods (all except linear regression) as described in Section~\ref{sec:bias}    \citep{correctBias}.

An overview of the results is given in \cref{tab:results} showing
the $R^2$, RMSE, and MAPE for the different algorithms on the test set, which was not used for model development.
Overall, the MSENet14 model achieved the best results with an $R^2$ of \num{0.823} for biomass (and carbon stocks)  and  an $R^2$ of \num{0.818} for wood volume. 
The corresponding RMSEs were \num{42.58} Mg\,ha$^{-1}$ and \num{81.45}  m$^3$\,ha$^{-1}$, respectively.

\begin{table}
    \centering
    \caption{Comparison of methods showing the $R^2$ score, RMSE, and MAPE on the test set. The $R^2$ and MAPE for biomass are the same as for carbon stocks. Note that we excluded 0 biomass measurements from MAPE to avoid numerical issues. The median (med.) and best result of 5 trials are shown, the best results are highlighted.}
    \label{tab:results}
    \begin{tabular}{
    ll
    S[table-format=1.3]
    S[table-format=1.3]
    S[table-format=3.2]
    S[table-format=2.2]
    S[table-format=4.2]
    S[table-format=3.2]
    }
        \toprule
     &                      & \multicolumn{2}{c}{$R^2$} & \multicolumn{2}{c}{RMSE} & \multicolumn{2}{c}{MAPE} \\
target & model   &    {med.} &    {max} &      {med.} &     {min} &       {med.} &       {min} \\
\midrule
AGB
     & linear &  0.757 & 0.757 &  49.88 & 49.88 &  577.78 & 577.78 \\
     & \power{} &  0.761 & 0.761 &  49.51 & 49.51 &  365.34 & 365.34 \\
     & RF &  0.751 & 0.752 &  50.47 & 50.45 &  950.49 & 931.58 \\
     & PointNet &  0.722 & 0.766 &  53.40 & 48.99 & 2039.97 & 997.48 \\
     & KPConv &  0.796 & 0.806 &  45.72 & 44.60 &  268.63 & 204.02 \\
     & MSENet14 &  \B 0.819 & \B 0.823 &  \B 43.00 & \B 42.58 &  230.36 & 102.16 \\
     & MSENet50 &  0.815 & 0.821 &  43.59 & 42.78 &  \B 127.33 &  \B 99.43 \\
    \addlinespace
volume
     & linear &  0.760 & 0.760 &  93.52 & 93.52 &  172.91 & 172.91 \\
     & \power{} &  0.763 & 0.763 &  92.82 & 92.82 &  223.64 & 223.64 \\
     & RF &  0.755 & 0.756 &  94.38 & 94.34 &  217.50 & 216.40 \\
     & PointNet &  0.727 & 0.772 &  99.64 & 91.09 &  470.42 & 262.44 \\
     & KPConv &  0.788 & 0.801 &  87.95 & 85.16 &   86.47 &  73.05 \\
     & MSENet14 &  \B 0.813 & \B 0.818 &  \B 82.55 & \B 81.45 &   70.05 &  \B 65.69 \\
     & MSENet50 &  0.810 & 0.816 &  83.16 & 81.84 &   \B 69.15 &  68.99 \\
         \bottomrule
    \end{tabular}
\end{table}

The RF performed the worst.  
The better performance of the linear regression with an $R^2$ of \num{0.757} for biomass and \num{0.760} for wood volume,  indicates that the calculated features  are to a certain extend linearly correlated with the target variables. 
The \power{} regression model based on \cite{magnussen2018lidar} reached $R^2$ values of \num{0.761} and \num{0.763} for biomass and tree volume, respectively, that is, the deep learning approach clearly improved over the state-of-the-art methods.
All more recent point cloud based methods outperformed the models based on the precomputed features, justifying that the networks indeed utilized additional information from the point clouds.
The MAPE, which was not an optimization criterion, showed the largest differences between the methods, which are mainly attributed to errors related to small target values.

\begin{figure}
    \centering
    \resizebox{\linewidth}{!}{\input{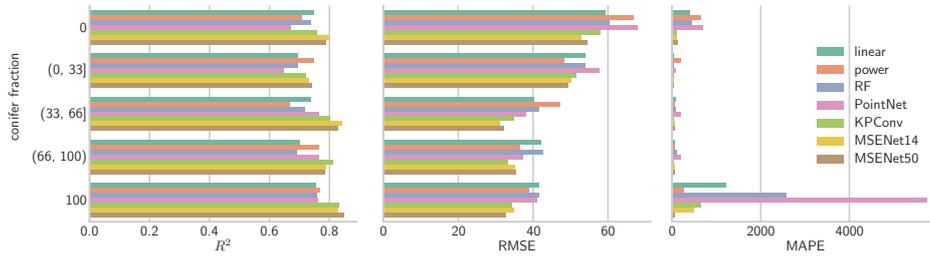}}
    \caption{Errors for different fractions of conifer and broadleaf trees for biomass on the test set. The columns show $R^2$ (higher is better), RMSE (lower is better), and MAPE (lower is better). 
    The corresponding figure for wood volume is qualitatively the same.
    \label{fig:species_bplot}}
\end{figure}

The national forest inventory data offers information about the fraction of conifer and broadleaf for each site, which we used to analyze the performance at different fractions in \cref{fig:species_bplot}. 
Interestingly, KPConv performed better than the Minkowski models for conifer fraction interval \((66, 100]\) and nearly equally good at 100.
Again, the high MAPE for the  baseline methods that rely on the precomputed features and PointNet was noticeable, in particular when mono cultures were analyzed. 
In general, mono cultures turned out to be more difficult than mixed forests w.r.t.\ MAPE but not w.r.t.\ RMSE and $R^2$ (e.g., forests with conifer trees only exhibited the lowest RMSE and $R^2$).

\begin{figure}
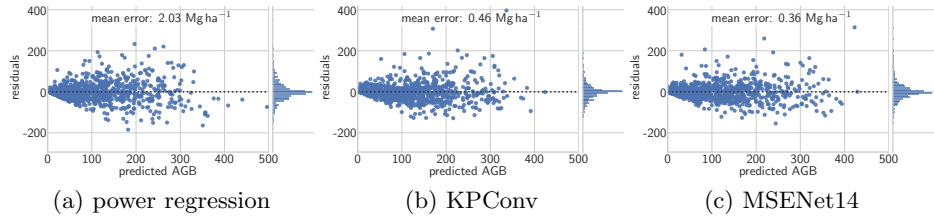

    \begin{subfigure}[b]{.33\linewidth}
    \resizebox{\linewidth}{!}{\input{fig/results_b/BMag_ha_power_resid_}}
    \caption{\power{} regression}
    \end{subfigure}~%
    \begin{subfigure}[b]{.33\linewidth}
    \resizebox{\linewidth}{!}{\input{fig/results_b/BMag_ha_KPConv_resid_}}
    \caption{KPConv}
    \end{subfigure}~%
    \begin{subfigure}[b]{.33\linewidth}
    \resizebox{\linewidth}{!}{\input{fig/results_b/BMag_ha_MSENet14_resid_}}
    \caption{MSENet14}
    \end{subfigure}
    \caption{Biomass residual (left side) and error distribution (right side) plots of the test performance for \power{} regression, KPConv, and MSENet14. The mean error is given as well to quantify the observed bias. %
    \label{fig:resid}}
\end{figure}

The spread of errors increased with magnitude of the target variables as seen in \cref{fig:resid} for the \power{}, KPConv, and MSENet14 model.
While low values were more often overestimated, high values tended to be underestimated.
Comparing the distribution of errors as well as the mean error of the assessed models, one can see that the KPConv and Minkowski models had a generally lower spread of errors than the \power{} model.
See \cref{fig:resid_b} and \ref{fig:resid_v} in the appendix for the remaining plots.

\begin{figure}
    \centering
    \resizebox{\linewidth}{!}{\input{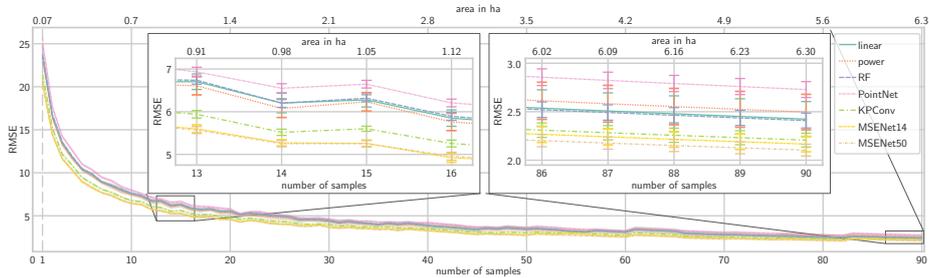}}
  \caption{Experiment showing  that errors cancel out if the area is increased. 
  The ordinate shows the carbon stock RMSE in \,Mg\,C\,ha$^{-1}$. 
  The abscissa indicates the number of randomly sampled subplots that are combined. Each subplot has a size 
of $\approx \SI{0.07}{ha}$, so 14 samples correspond to little less than $\SI{1}{ha}$.
The error bars indicate the standard error over 10 different random combinations of subplots for each model repetition.}
  \label{fig:spatial_agg_C}
\end{figure}

It has been found that the errors of ALS-based carbon stock estimates decrease when increasing the size of the observed area (e.g., aggregating over a larger area decreases the error~\citep{asner2010high,mascaro2011evaluating,dalponte2016tree,ferraz2016lidar}).
To demonstrate this, we calculated the RMSE for each model over the same randomly chosen combination of subplots (\cref{fig:spatial_agg_C}).
We repeated that process 10 times with different random combinations for each model repetition.
The predictions for the combined plots were added and compared to the aggregated measurements. 
We consider the carbon stock RMSE in \,Mg\,C\,ha$^{-1}$ to allow for easier comparison with published results.
The RMSE reduces quickly for all models and the ranking of models remains the same.

\section{Discussion}\label{sec:discussion}

The results give clear evidence that the deep learning approaches KPConv and MSENet are preferable to the methods working on point cloud statistics.
As the time to apply and train MSENet is drastically shorter (see \cref{tab:params_time}) and the results were marginally better, we regard MSENet as the method of choice.
We assume that the subpar performance of PointNet is due to the missing local context processing in the algorithm, which seems to be required for forest scenes.

\subsection{Quantitative comparison to other studies}
Caution is warranted when comparing  our prediction accuracy with that of other studies.
Reported errors of ALS-based prediction models for AGB range from \num{17} to \num{40}\,Mg\,ha$^{-1}$ in the tropics \citep{asner2011high}, and this error is comparable to estimates reported in other regions \citep{lefsky2002lidar} as well as previous studies with the Danish NFI~\citep{NORDLARSEN2012}. 
Reassuringly, we achieve error rates at the lower end of this spectrum. 
It is important to emphasize, however, that it is not uncommon that the indicators documented in these studies measure the error on a training data set as a goodness of fit index (\citealp[e.g., in the study by ][]{LEFSKY199983}  the $R^2$ for AGB drops from 80\,\% down to 33\,\% when the model is applied to a different data set).
Similarly but not as severe, earlier studies of the Danish NFI with fewer data also reported a \num{0.6}\% increase in RMSE when using hold-out sets~\citep{NORDLARSEN2012}.
Conversely, the training errors of our models are similar to the test set results in \cref{tab:results} (see \cref{tab:results_train} in the supplementary material), indicating that the models are not overfitted.

The error of the overall carbon stock prediction has been shown to decrease with increasing area, which is explained by error cancellation
~\citep{asner2010high,mascaro2011evaluating,dalponte2016tree,ferraz2016lidar,magnussen2018lidar}.
This is an important property when it comes to country-wide quantification of carbon.
For example, in the study by \cite{mascaro2011evaluating},
the RMSE of the estimated carbon stocks decreased from \num{63.2} to \num{11.1} and then to \num{6.5} \,Mg\,C\,ha$^{-1}$ when the observed area was increased from \num{0.04} to \num{1} and then to \SI{6.25}{ha}.
Our study areas are of size \(a=15^2\cdot\pi \cdot 10^{-4}\,\text{ha} \approx 0.07\,\text{ha}\). 
The residuals in \cref{fig:resid} indeed suggest that our errors  average out if several subplots are combined.
This is confirmed by the results shown in \cref{fig:spatial_agg_C}.
For aggregated areas of approximately \SI{1}{ha} (with 14 samples it is \SI{0.98}{ha}) and \SI{6.25}{ha} (with 89 samples it is \SI{6.23}{ha}), the RMSE for the MSENet14 model was \num{5.28} (SE \num{0.09}) and \num{2.19} (SE \num{0.09}), respectively. 
Even though the errors cancel out, a good high-resolution performance seems to translate to a better aggregated performance. %

\subsection{Applicability and map generation}
The results are evaluated on an previously unseen and non-overlapping test set, meaning that the model should exhibit a similar performance on other arbitrarily chosen forest subplots from the study area (e.g., Denmark).
For new study areas that are dissimilar to ours (e.g., different climate zone, tree species, or topographical features), reevaluation and potential fine-tuning of the models would be needed.
However, since little preprocessing is required, this verification and adjustment step is comparatively minor.

\begin{figure}
    \centering
    \resizebox{.99\linewidth}{!}{\input{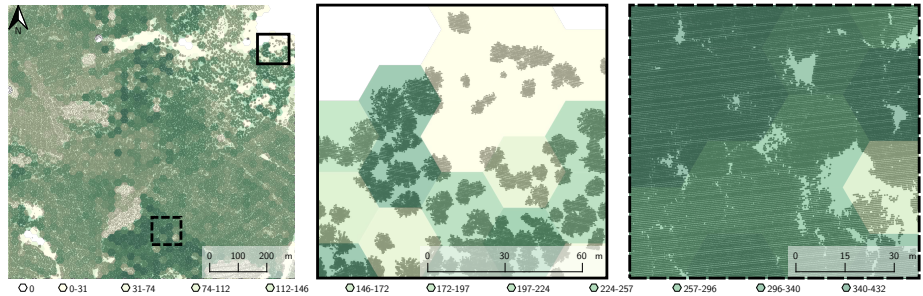}}
    \caption{%
    Exemplary map showing predictions at different zoom levels. 
    Points of the point cloud colored grey (only $\approx \SI{1}{\metre}$ above the ground) and overlaid with the estimated biomass in \,Mg\,ha$^{-1}$, where a deeper green indicates more biomass.
    \label{fig:map_pred}}
\end{figure}

Our method allows for dense, non-interpolated biomass mappings with unprecedented accuracy at large scale at a resolution below \SI{30}{\metre}. %
An exemplary biomass map is demonstrated in \cref{fig:map_pred}.
The figure shows the results of the MSENet50 model applied to a randomly selected \SI{1}{\kilo\metre} area in Denmark without any additional preprocessing.
The results are convincing, as areas with no points above ground are predicted to have no biomass, and areas with many trees have high biomass.
Thus, our method enables highly accurate analyses of the AGB development in Denmark by comparing  AGB maps from different years.

\section{Conclusions}\label{sec:conclusions}
Precise quantification of the spatial distribution of forest biomass and carbon content is a valuable tool to understand the processes driving {af-}, re-, and deforestation, and the variations in carbon cycle from the ecosystem to the regional and global scale~\citep{zolkos2013meta}.

This study brings forward deep learning systems for predicting above-ground biomass, wood volume, and carbon stocks in forests efficiently and directly from airborne \lidar{} point clouds.
Three conceptually different neural network architectures for classification (i.e., PointNet, KPConv, and  Minkowski CNNs) were modified to perform regression, and were evaluated using a unique data set combining field measurements and \lidar{} data.

Our adaptation of Minkowski CNNs outperformed the other deep learning approaches as well as the baseline methods. 
Specifically, the coefficient of determination $R^2$ for above-ground biomass and stored carbon exceeded \num{0.82} at \SI{0.07}{ha} on a test set, which was not considered in the modelling process. 
The main advantage of the method proposed is the efficient implementation of 3D convolutions for sparse data, as it relies on the same operations that CNNs use for standard image processing.

The assessed deep learning models achieved high accuracies, leading to a considerably better performance in predicting AGB of forest areas, compared to the state-of-the-art approach operating on basic statistics of the point clouds.
Our approach 
simplifies the production of AGB and carbon content maps at high resolution. 
It does neither require  digital elevation models (e.g., DTMs or canopy height models) nor forest canopies with structural heterogeneity.

These encouraging results may be the first step towards a paradigm shift in utilizing \lidar{} data for accurate quantification and analysis of forest carbon dynamics, which is a key component for achieving the United Nations Strategic Plan for Forests under the Paris agreement and the Sustainable Development Goals (SDGs)~\citep{forestgoals,sdg}  in accordance with
the IPCC 2018~\citep{ipcc2018}).

\section*{Acknowledgement}
This work was supported by the
research grant DeReEco (34306)  from Villum Foundations, the Independent Research Fund Denmark through the grant \emph{Monitoring Changes in Big Satellite Data via Massively-Parallel Artificial Intelligence} (9131-00110B), a Villum Experiment grant by the Velux Foundations (MapCland project, project number: 00028314), the DeepCrop project (UCPH Strategic plan 2023 Data+ Pool), and the Pioneer Centre for AI, DNRF grant number P1.

\bibliographystyle{plainnat}
\bibliography{macros,pointcloud}
\clearpage

\appendix

\section{Additional Results}
\begin{figure}
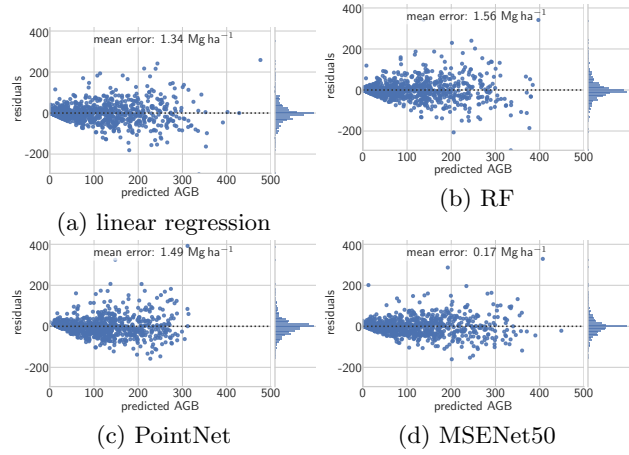

    \centering
    \begin{subfigure}[b]{.33\linewidth}
    \resizebox{\linewidth}{!}{\input{fig/results_b/BMag_ha_linear_resid_}}
    \caption{linear regression}
    \end{subfigure}~%
    \begin{subfigure}[b]{.33\linewidth}
    \resizebox{\linewidth}{!}{\input{fig/results_b/BMag_ha_RF_resid_}}
    \caption{RF}~%
    \end{subfigure}
    \begin{subfigure}[b]{.33\linewidth}
    \resizebox{\linewidth}{!}{\input{fig/results_b/BMag_ha_PointNet_resid_}}
    \caption{PointNet}
    \end{subfigure}
    \begin{subfigure}[b]{.33\linewidth}
    \resizebox{\linewidth}{!}{\input{fig/results_b/BMag_ha_MSENet50_resid_}}
    \caption{MSENet50}
    \end{subfigure}
    \caption{Biomass residual (left side) and error distribution (right side) plots of the test performance.\label{fig:resid_b}}
\end{figure}

\begin{figure*}[ht]
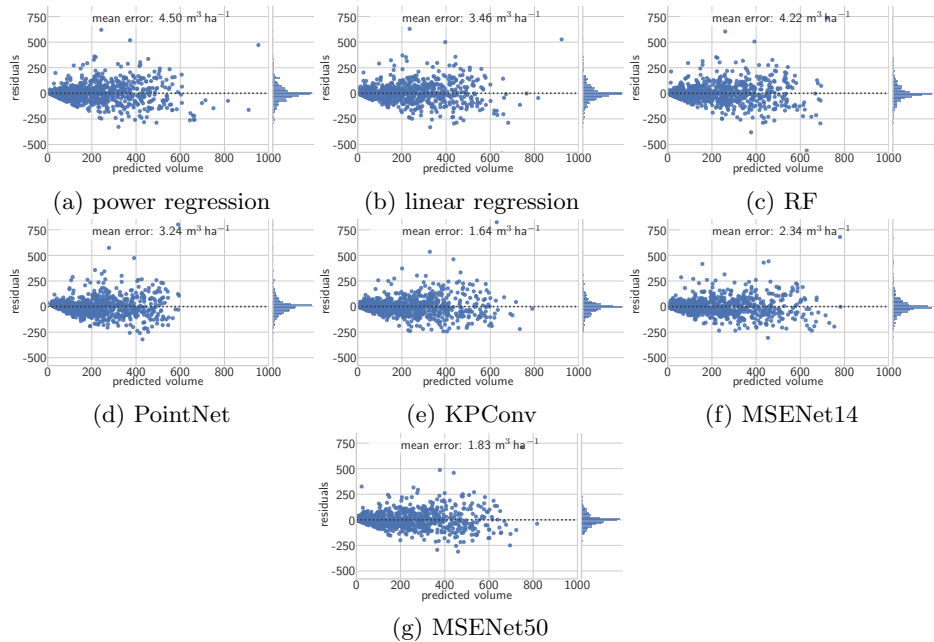

    \centering
    \begin{subfigure}[b]{.33\linewidth}
    \resizebox{\linewidth}{!}{\input{fig/results_v/V_ha_power_resid_}}
    \caption{\power{} regression}
    \end{subfigure}~%
    \begin{subfigure}[b]{.33\linewidth}
    \resizebox{\linewidth}{!}{\input{fig/results_v/V_ha_linear_resid_}}
    \caption{linear regression}
    \end{subfigure}~%
    \begin{subfigure}[b]{.33\linewidth}
    \resizebox{\linewidth}{!}{\input{fig/results_v/V_ha_RF_resid_}}
    \caption{RF}
    \end{subfigure}
    \begin{subfigure}[b]{.33\linewidth}
    \resizebox{\linewidth}{!}{\input{fig/results_v/V_ha_PointNet_resid_}}
    \caption{PointNet}
    \end{subfigure}~%
    \begin{subfigure}[b]{.33\linewidth}
    \resizebox{\linewidth}{!}{\input{fig/results_v/V_ha_KPConv_resid_}}
    \caption{KPConv}
    \end{subfigure}~%
    \begin{subfigure}[b]{.33\linewidth}
    \resizebox{\linewidth}{!}{\input{fig/results_v/V_ha_MSENet14_resid_}}
    \caption{MSENet14}
    \end{subfigure}
    \begin{subfigure}[b]{.33\linewidth}
    \resizebox{\linewidth}{!}{\input{fig/results_v/V_ha_MSENet50_resid_}}
    \caption{MSENet50}
    \end{subfigure}
    \caption{Wood volume residual (left side) and error distribution (right side) plots of the test performance.\label{fig:resid_v}}
\end{figure*}

\cref{tab:results_train} and \cref{tab:results_val} presents the results on the training and validation set, respectively.
Note that the models using extracted features  (linear regression, \power{} regression, and RF) were trained on the union of training and validation set, while our point cloud regression models were not.
Still, the more recent models directly working on the point cloud data consistently produced better results.
\cref{fig:resid_b} and \ref{fig:resid_v} illustrate residual plots and error distribution of biomass and wood volume, respectively, which completes \cref{fig:resid}.

\begin{figure*}[ht]
    \centering
     \begin{subfigure}[b]{\linewidth}
         \centering
        \includegraphics[width=.8\linewidth]{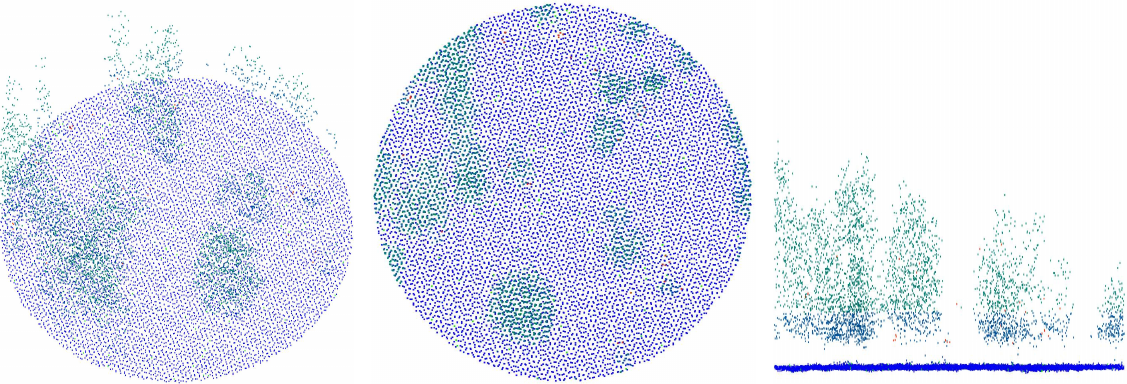}
        \caption{Target: \num{1.79}; MSENet14: \num{1.59}; KPConv: \num{3.16}; PointNet: \num{10.17}; RF: \num{8.70}, lin.\ model: \num{10.42}; \power{} model: \num{11.31}}
    \end{subfigure}
    
     \begin{subfigure}[b]{\linewidth}
         \centering
        \includegraphics[width=.8\linewidth]{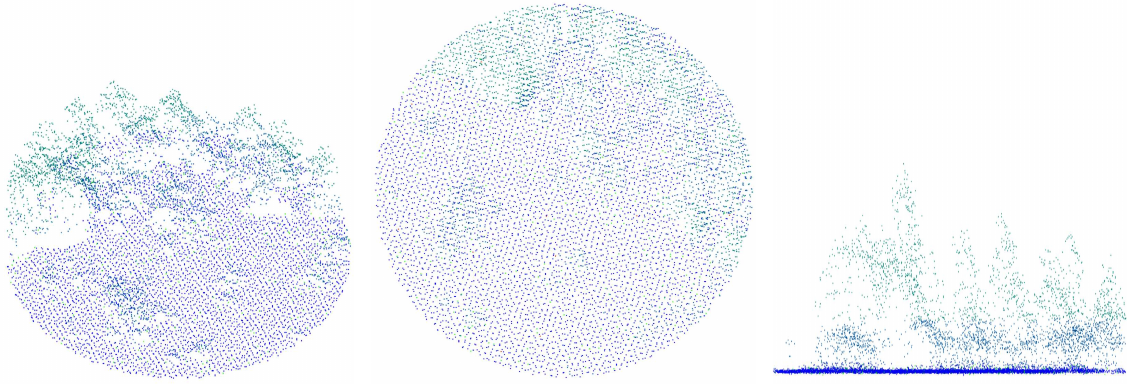}
        \caption{Target: \num{0.86}; MSENet14: \num{1.00}; KPConv: \num{3.99}; PointNet: \num{11.31}; RF: \num{7.00}, lin.\ model: \num{8.09}; \power{} model: \num{9.94}}
    \end{subfigure}

    \begin{subfigure}[b]{\linewidth}
        \centering
        \includegraphics[width=.8\linewidth]{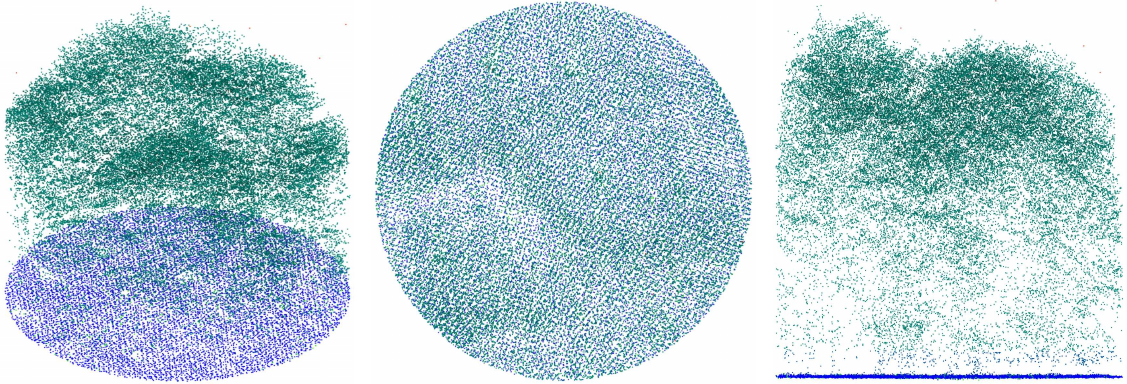} 
      \caption{Target: \num{417.60}; MSENet14: \num{345.71}; KPConv: \num{381.54}; PointNet: \num{298.25}; RF: \num{285.75}, lin.\ model: \num{272.74}; \power{} model: \num{270.98}}
    \end{subfigure}

    \begin{subfigure}[b]{\linewidth}
        \centering
    \includegraphics[width=.8\linewidth]{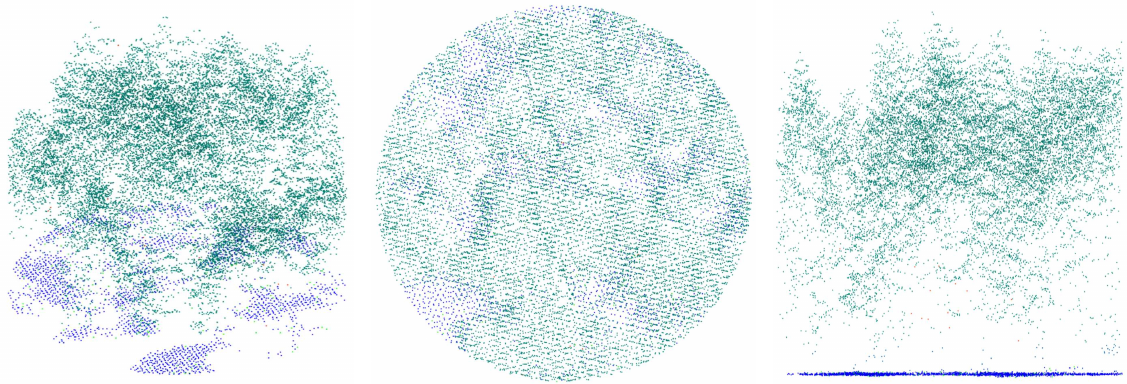}
      \caption{Target: \num{245.90}; MSENet14: \num{212.94}; KPConv: \num{273.01}; PointNet: \num{195.50}; RF: \num{216.13}, lin.\ model: \num{212.38}; \power{} model: \num{217.86}}
    \end{subfigure}

    \begin{subfigure}[b]{\linewidth}
        \centering
        \includegraphics[width=.8\linewidth]{fig/sample/Position.pdf}
    \end{subfigure}

    \caption{
    More examples of subplots in each row with three perspectives: isometric front, top, and side view. Each subplots caption contains the measured and estimated biomass in Mg\,ha$^{-1}$.}
    \label{fig:samples-more}
\end{figure*}

\begin{figure*}
    \centering
        \includegraphics[width=\linewidth]{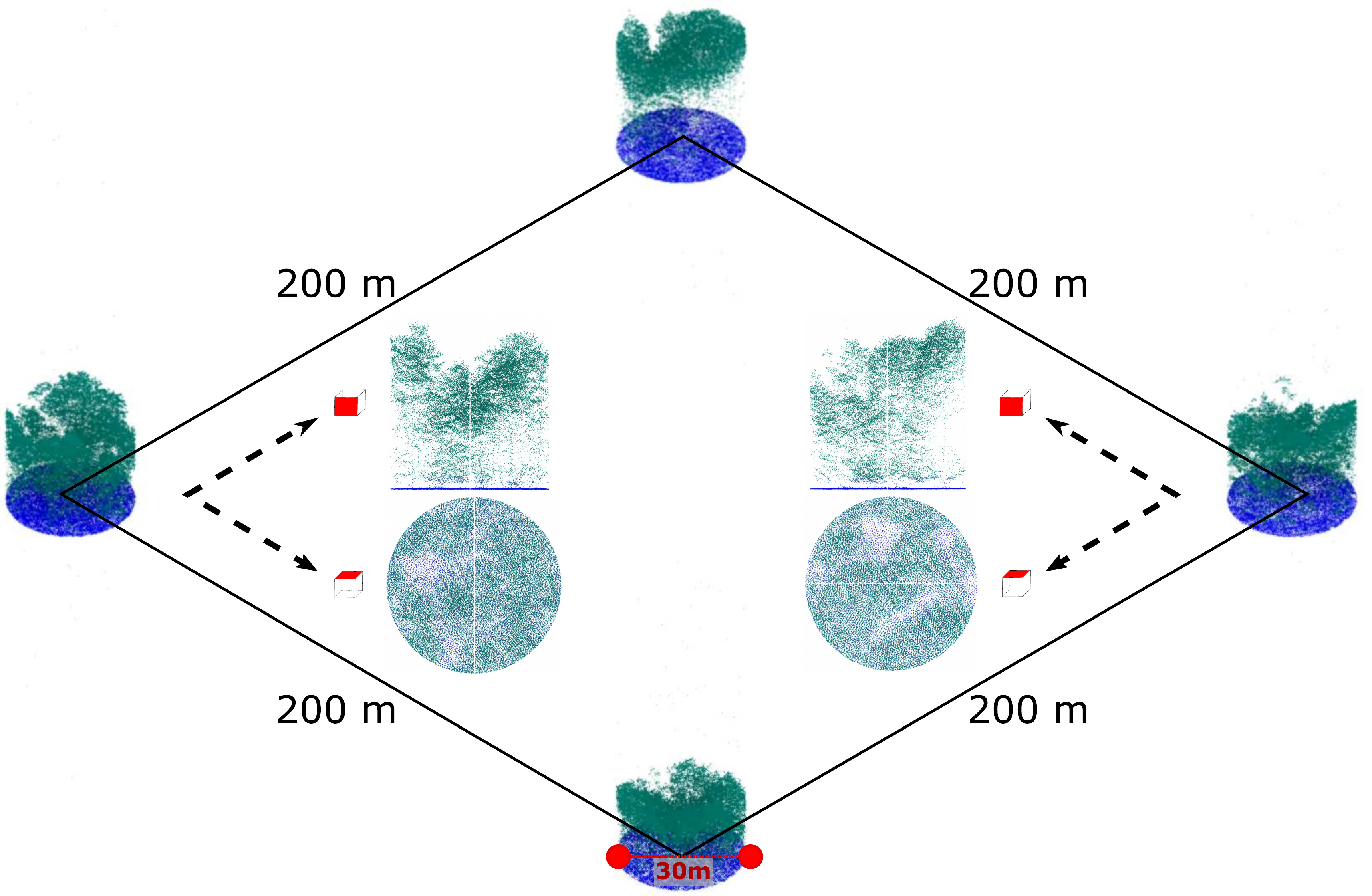}
    \caption{One plot sample from a $2\times\SI{2}{\kilo\metre}$ area as described in \cref{sec:forest_data}. Each plot usually contains four subplots in a $200 \times \SI{200}{\metre}$ square. All \lidar{} subplots have a diameter of \SI{30}{\metre} and their own biomass as well as wood volume measurements.
    } \label{fig:plot-samples}
\end{figure*}

In \cref{fig:samples-more} we show more subplots with particularly low or high values for above-ground biomass.
For example, the first two subplots have low AGB with \num{1.79} and \num{1.89}\,Mg\,ha$^{-1}$, which is reflected by the sparse vegetation in the point cloud.
This is difficult to capture with commonly used statistics derived from point cloud data, and therefore models based on these features (e.g., RF, lin., and power model) overestimated.
Interestingly, PointNet also overestimate values due to the employed global aggregation without intermediate downsampling steps (similar to global feature extraction).
The third example in \cref{fig:samples-more} exhibits a high AGB value with \num{417.60}\,Mg\,ha$^{-1}$ and the point cloud shows a densely populated and natural broadleaf forest.
Again, this is difficult to estimate since the derived statistics do not offer information about individual biomass contributions and thus most methods underestimate them.
The fourth subplots biomass is \num{245.89}\,Mg\,ha$^{-1}$ from a conifer forest and the predictions align well with the target.
We also give a visualization of the plots and subplots in \cref{fig:plot-samples} (as described in \cref{sec:forest_data}).

\begin{table}
    \centering
    \caption{Comparison of methods showing the $R^2$ score, RMSE, and MAPE on the training set. The models with best validation set performances were chosen.  The $R^2$ and MAPE for biomass are the same as for carbon stocks. Note that we excluded 0 biomass measurements from MAPE to avoid numerical issues. Best results are highlighted.}
    \label{tab:results_train}
    \begin{tabular}{
    ll
    S[table-format=1.3]
    S[table-format=1.3]
    S[table-format=3.2]
    S[table-format=3.2]
    S[table-format=4.2]
    S[table-format=4.2]
    }
        \toprule
     &                      & \multicolumn{2}{c}{$R^2$} & \multicolumn{2}{c}{RMSE} & \multicolumn{2}{c}{MAPE} \\
target & model   &    {med.} &    {max} &      {med.} &     {min} &       {med.} &       {min} \\
\midrule
AGB  
     & linear &  0.712 & 0.712 &  55.47 &  55.47 &  770.03 &  770.03 \\
     & \power{} &  0.700 & 0.700 &  56.63 &  56.63 &  690.59 &  690.59 \\
     & RF &  0.759 & 0.760 &  50.75 &  50.73 &  553.61 &  548.55 \\
     & PointNet &  0.684 & 0.706 &  58.16 &  56.13 & 2146.63 & 1273.65 \\
     & KPConv &  \B 0.795 & \B 0.799 & \B 46.84 & \B 46.37 &  267.56 & \B 129.17 \\
     & MSENet14 &  0.786 & 0.794 &  47.81 &  46.91 &  318.26 &  259.58 \\
     & MSENet50 &  0.791 & 0.800 &  47.32 &  46.22 & \B 250.04 &  184.65 \\

\addlinespace
volume 
     & linear &  0.716 & 0.716 & 103.51 & 103.51 &  188.70 &  188.70 \\
     & \power{} &  0.703 & 0.703 & 105.83 & 105.83 &  197.08 &  197.08 \\
     & RF &  0.765 & 0.765 &  94.15 &  94.11 &  157.32 &  156.09 \\
     & PointNet &  0.694 & 0.714 & 107.39 & 103.92 &  431.84 &  283.08 \\
     & KPConv &  0.792 & 0.798 &  88.58 &  87.38 &   91.97 &  \B 68.84 \\
     & MSENet14 &  0.784 & 0.792 &  90.27 &  88.54 &   96.16 &   81.57 \\
     & MSENet50 & \B 0.794 & \B 0.801 &  \B 88.19 & \B 86.57 &  \B 87.61 &   85.31 \\

    \bottomrule
    \end{tabular}
\end{table}

\begin{table}
    \centering
    \caption{Comparison of methods showing the $R^2$ score, RMSE, and MAPE on the validation set. The $R^2$ and MAPE for biomass are the same as for carbon stocks. Note that we excluded 0 biomass measurements from MAPE to avoid numerical issues. Best results are highlighted. Note that the deep learning models did not train on this data directly but the feature extraction methods did.}
    \label{tab:results_val}
    \begin{tabular}{
    ll
    S[table-format=1.3]
    S[table-format=1.3]
    S[table-format=3.2]
    S[table-format=2.2]
    S[table-format=4.2]
    S[table-format=4.2]
    }
        \toprule
     &                      & \multicolumn{2}{c}{$R^2$} & \multicolumn{2}{c}{RMSE} & \multicolumn{2}{c}{MAPE} \\
target & model   &    {med.} &    {max} &      {med.} &     {min} &       {med.} &       {min} \\
\midrule
AGB  
     & linear &  0.736 & 0.736 &  54.37 & 54.37 &  388.13 &  388.13 \\
     & \power{} &  0.739 & 0.739 &  53.99 & 53.99 &  384.63 &  384.63 \\
     & RF &  0.775 & 0.775 &  50.16 & 50.12 &  398.48 &  375.35 \\
     & PointNet &  0.685 & 0.728 &  59.32 & 55.21 & 1811.21 & 1303.69 \\
     & KPConv &  0.773 & 0.775 &  50.37 & 50.19 &  176.69 &  144.17 \\
     & MSENet14 &  0.796 & 0.798 &  47.77 & 47.49 &  207.74 &   92.94 \\
     & MSENet50 &  \B 0.805 & \B 0.810 & \B 46.70 & \B 46.08 & \B  71.72 & \B  61.81 \\

\addlinespace
volume 
     & linear &  0.764 & 0.764 &  95.05 & 95.05 &  114.59 &  114.59 \\
     & \power{} &  0.766 & 0.766 &  94.61 & 94.61 &  118.03 &  118.03 \\
     & RF &  0.800 & 0.801 &  87.45 & 87.40 &  111.73 &  109.71 \\
     & PointNet &  0.719 & 0.760 & 103.81 & 95.94 &  322.45 &  205.49 \\
     & KPConv &  0.794 & 0.800 &  88.87 & 87.61 &   72.38 &   65.82 \\
     & MSENet14 &  0.814 & 0.816 &  84.44 & 83.95 &   69.24 & \B  60.66 \\
     & MSENet50 &  \B 0.824 & \B 0.826 &  \B 82.20 & \B 81.53 & \B  61.76 &   61.22 \\
    \bottomrule
    \end{tabular}
\end{table}

 \begin{figure}[t]
 \centering
 \resizebox{\linewidth}{!}{\input{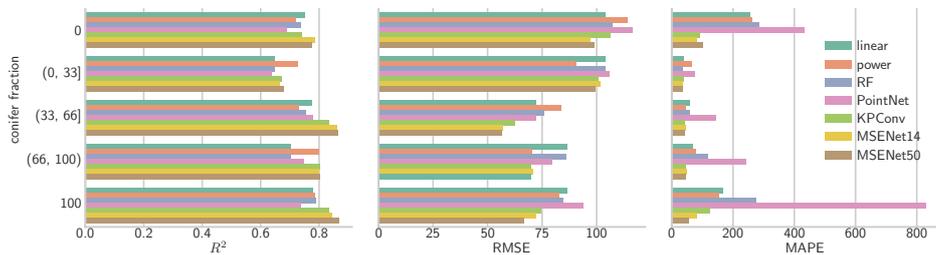}}
 \caption{Errors for different fractions of conifer and broadleaf trees for wood volume on the test set. The columns show $R^2$ (higher is better), RMSE (lower is better), and MAPE (lower is better).\label{fig:species_bplot_appendix}}
\end{figure}


\end{document}